\documentclass[10pt,twocolumn,letterpaper]{article}

\usepackage[pagenumbers]{wacv} 

\usepackage{graphicx}
\usepackage{amsmath}
\usepackage{amssymb}
\usepackage{booktabs}
\usepackage{multirow}

\usepackage[pagebackref,breaklinks,colorlinks]{hyperref}

\usepackage[capitalize]{cleveref}

\begin{document}

\title{RopeTP: Global Human Motion Recovery via Integrating Robust Pose Estimation with Diffusion Trajectory Prior}

\author{
    \begin{minipage}{0.2\textwidth}
        \centering
        Mingjiang Liang*\\
    \end{minipage}
    \hfill
    \begin{minipage}{0.2\textwidth}
        \centering
        Yongkang Cheng\thanks{means these authors contributed equally} \\
    \end{minipage}
    \hfill
    \begin{minipage}{0.2\textwidth}
        \centering
        Hualin Liang \\
    \end{minipage}
    \hfill
    \begin{minipage}{0.2\textwidth}
        \centering
        Shaoli Huang\thanks{is corresponding author.} \\
    \end{minipage}
    \begin{minipage}{0.1\textwidth}
        \centering
        Wei Liu \\
    \end{minipage}
}
\maketitle
\begin{abstract}
   We present RopeTP, a novel framework that combines \textbf{Ro}bust \textbf{p}ose \textbf{e}stimation with a diffusion \textbf{T}rajectory \textbf{P}rior to reconstruct global human motion from videos. At the heart of RopeTP is a hierarchical attention mechanism that significantly improves context awareness, which is essential for accurately inferring the posture of occluded body parts. This is achieved by exploiting the relationships with visible anatomical structures, enhancing the accuracy of local pose estimations. The improved robustness of these local estimations allows for the reconstruction of precise and stable global trajectories. Additionally, RopeTP incorporates a diffusion trajectory model that predicts realistic human motion from local pose sequences. This model ensures that the generated trajectories are not only consistent with observed local actions but also unfold naturally over time, thereby improving the realism and stability of 3D human motion reconstruction. Extensive experimental validation shows that RopeTP surpasses current methods on two benchmark datasets, particularly excelling in scenarios with occlusions. It also outperforms methods that rely on SLAM for initial camera estimates and extensive optimization, delivering more accurate and realistic trajectories.
\end{abstract}

\begin{figure}[t]
  \begin{center}
    \includegraphics[width=1.0\linewidth]{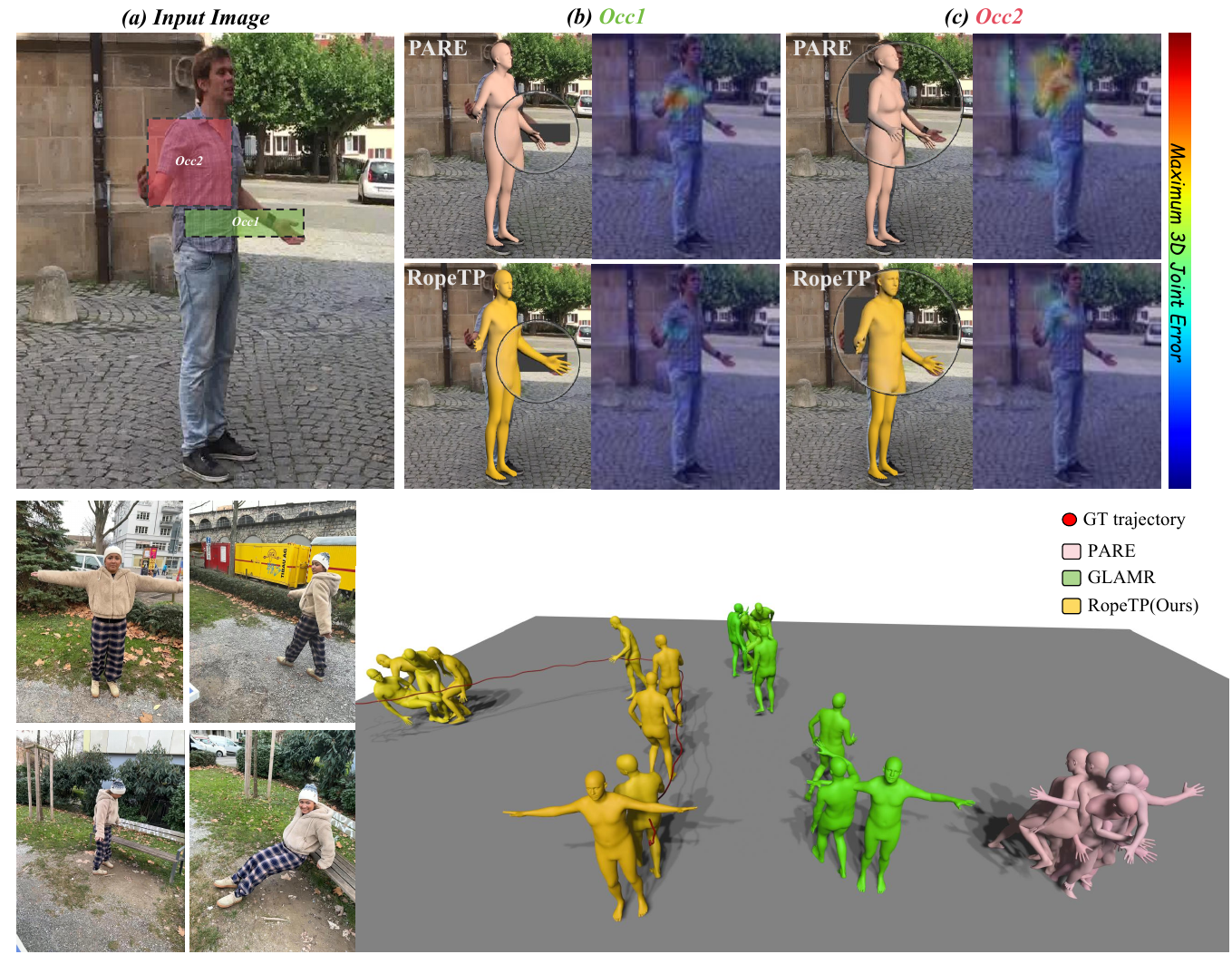}
  \end{center}
  \caption{The upper part of the figure compares the occlusion handling in human shape recovery between the PARE method and our proposed method. Panel (a) displays the original image with two different shaped occluders. Panels (b) and (c) show the estimation results of PARE and our method under the two occlusion scenarios. The occlusion sensitivity map on the far right quantifies the maximum joint estimation error for each pixel position of the occluder, highlighting the robustness of the proposed method against various shape occlusions. Unlike PARE, which lacks global trajectory information, our method is capable of regenerating reasonable global trajectories while reconstructing robust poses. As illustrated in the lower half of the figure, the robust poses provide a powerful prior for trajectory generation, resulting in human motion trajectories that are nearly consistent with the video.}
  \vspace{-2em}
  \label{fig:title_img}
  \end{figure}
  
\section{Introduction}
\label{sec:intro}

Imagine how you reason about missing pieces while solving a jigsaw puzzle or how you infer a person's motion trajectory from a single picture. This is analogous to the challenge in computer vision of estimating human shape and pose from images captured by a single camera. Just as you can use surrounding puzzle pieces to deduce the appearance of the missing piece (or exploit the joint positions of human motion to infer their trajectory), our method aims to reconstruct a complete 3D human mesh and precise motion trajectories from images, even when certain body parts are occluded or obscured.

In the past, the field heavily relied on parametric models such as SMPL~\cite{loper2015smpl}, which transformed the complex task of 3D mesh reconstruction into a more manageable parametric regression problem. However, regression-based methods~\cite{kanazawa2018end, Moon2020i2l,kolotouros2019learning} often fail when encountering occlusions, much like how missing puzzle pieces make it difficult to perceive the whole picture.

This challenge becomes even more pronounced when body parts are occluded by themselves or external objects, a situation frequently encountered in the real world. Previous methods~\cite{kanazawa2018end, kolotouros2019learning} struggled with this, as they were often overly sensitive to such occlusions. Recent approaches~\cite{kocabas2021pare, xue20223d} attempted to address this issue by focusing on individual body parts, but these too had their limitations, typically relying excessively on the limited visible information surrounding the occluded areas. Consequently, these methods remain highly sensitive to occlusion issues, as illustrated in the upper half of Figure~\ref{fig:title_img}. 

Our work is inspired by the human ability to infer hidden details using contextual cues. Just as you might guess the content of a missing puzzle piece based on adjacent pieces, our method leverages multi-scale visual cues to reconstruct occluded body parts. We introduce a novel architecture that combines a hierarchical attention-guided segmenter with an adaptive contextual part regressor. This framework excels at extracting and synthesizing multi-scale visual information, much like assembling a puzzle while considering both close-up details and the bigger picture. Moreover, monocular camera-based reconstruction methods have long been plagued by the ambiguity of human trajectories, leading to a sharp decline in performance in 3D space. We emulate the human process of inferring motion trajectories based on joint poses by introducing a diffusion generative model, which re-infers global motion trajectories conditioned on the reconstructed joint poses to rectify the ambiguous global information. It's worth noting that this approach is efficient, which distinguishes it from the time-consuming trajectory optimization strategies typically employed in dynamic camera systems~\cite{yuan2022glamr, sun2023trace}. Simultaneously, accurate human pose estimation, as a powerful prior knowledge, can assist the diffusion model in regenerating precise human trajectories. As depicted in the lower half of Figure~\ref{fig:title_img}, our method achieves a significant lead over both the static PARE~\cite{kocabas2021pare} and dynamic GLAMR~\cite{yuan2022glamr} approaches.

Our method demonstrates exceptional performance on standard datasets such as 3DPW~\cite{von2018recovering} and Human3.6M~\cite{ionescu2013human3}, as well as on occlusion-specific datasets, showcasing its robustness in handling occlusion complexity. In addition, we validate the precision of our global robust human mesh estimation by comparing it with dynamic camera methods on the EMDB~\cite{kaufmann2023emdb} dataset. Extensive qualitative analyses, ablation studies, and video results in the supplementary material underscore the efficacy of our approach in resolving monocular camera trajectory ambiguity and addressing occlusion challenges.

\section{Related Work}
\noindent \textbf{Regression-based methods.} 
Regressing human pose and shape parameters from a single RGB image can be categorized into two main approaches: auxiliary regression and direct regression. 
Auxiliary regression strategies leverage prior knowledge from ancillary domains to facilitate parameter regression. Methods such as HybrIK~\cite{li2021hybrik} and IKOL~\cite{zhang2023ikol} decompose pose parameters into a two-step motion process, while PARE~\cite{kocabas2021pare} and A-LS~\cite{xue20223d} predict attention masks for regression. On the other hand, direct regression approaches extract features from a single image for model parameter regression without relying on auxiliary information. HMR~\cite{kanazawa2018end} employs a CNN network and MLP layers for this purpose, while ROMP~\cite{sun2021monocular} maps each 3D pixel channel onto model parameter channels. Building upon HMR, CLIFF~\cite{li2022cliff} utilizes bounding box information and introduces a full-image projection to enhance prediction results. Despite significant advancements in regression-based methods for human pose and shape estimation, their ability to predict global motion trajectories remains limited, constraining their practical applicability in real-world situations.

\noindent \textbf{Occlusion handling.} Addressing occlusion is crucial in 3D human reconstruction. Acquiring occluded data is challenging, leading to artificial occlusions in previous studies~\cite{biggs20203d, rockwell2020full, georgakis2020hierarchical, cheng20203d}. However, this often leads to unsatisfactory real-world outcomes.

Inferring occluded portions involves establishing relationships between discernible cues. VisDB~\cite{yao2022learning} predicts mesh vertex coordinates and visibility labels. PARE~\cite{kocabas2021pare} and A-LS~\cite{xue20223d} leverage implicit spatial relationships and adaptive attention mechanisms, respectively. Despite their ability to recover poses obscured by small-scale occlusions, errors involving part depth confusion often emerge in self-occlusion scenarios. BoPR~\cite{cheng2023bopr} proposes a body-aware reference-based approach to combat depth ambiguity. However, occlusion remains an ill-posed problem. In contrast, our method harnesses attention mechanisms across diverse scales to guide the aggregation of multiple body-aware features, generating plausible results.

\noindent \textbf{Global human motion reconstruction.} 
The task of deriving global human trajectories from a monocular dynamic camera is a complex one. Traditional methods have depended on previously learned human motion distributions to differentiate between human and camera movements. For example, GLAMR~\cite{yuan2022glamr} forecasts the global trajectory by leveraging a predicted and infilled 3D motion sequence, optimizing it for multiple individuals in a given scene. However, the lack of consideration for camera motion cues in GLAMR can result in noisy trajectories, particularly when the camera is rotating. Alternative techniques like SLAHMR~\cite{ye2023decoupling} and PACE~\cite{bartoli2017pace} utilize readily available SLAM algorithms to simultaneously optimize both camera and human movements with the aim of minimizing the negative log likelihood of a learned motion prior~\cite{rempe2021humor}. While these traditional methods can produce satisfactory results, their optimization process is computationally intensive, potentially limiting their practical application. In contrast, our proposed method employs a robust pose estimator and learns trajectory prior using diffusion models, providing a more efficient and accurate solution for predicting human trajectories.

\noindent\textbf{Preliminaries:}
In this section, we provide a brief overview of the human parametric model SMPL, which our work is based upon. \\
The SMPL model is a widely utilized parametric model for human body shape and pose estimation. Our method employs the SMPL parametric model to represent the human body. The model necessitates two essential parameters: pose, denoted as $\theta \in {\mathbb{R}}^{72}$, and shape, denoted as $\beta \in {\mathbb{R}}^{10}$.The SMPL model generates a 3D mesh $\mathcal{M}(\theta,\beta)\in {\mathbb{R}}^{6890\times3}$ that represents the human body in a differentiable manner. Specifically, the mesh is defined by a set of vertices that are positioned in 3D space according to the pose and shape parameters. The reconstructed 3D joints are obtained by applying a pre-trained linear regression matrix $\mathcal{W}$ to the mesh vertices, resulting in $\mathcal{J}_{3D}=\mathcal{W} \mathcal{M} \in {\mathbb{R}}^{J\times 3}$, where $J=24$. 

\section{Method}
\begin{figure}[t]
  \begin{center}
    \includegraphics[width=0.75\linewidth]{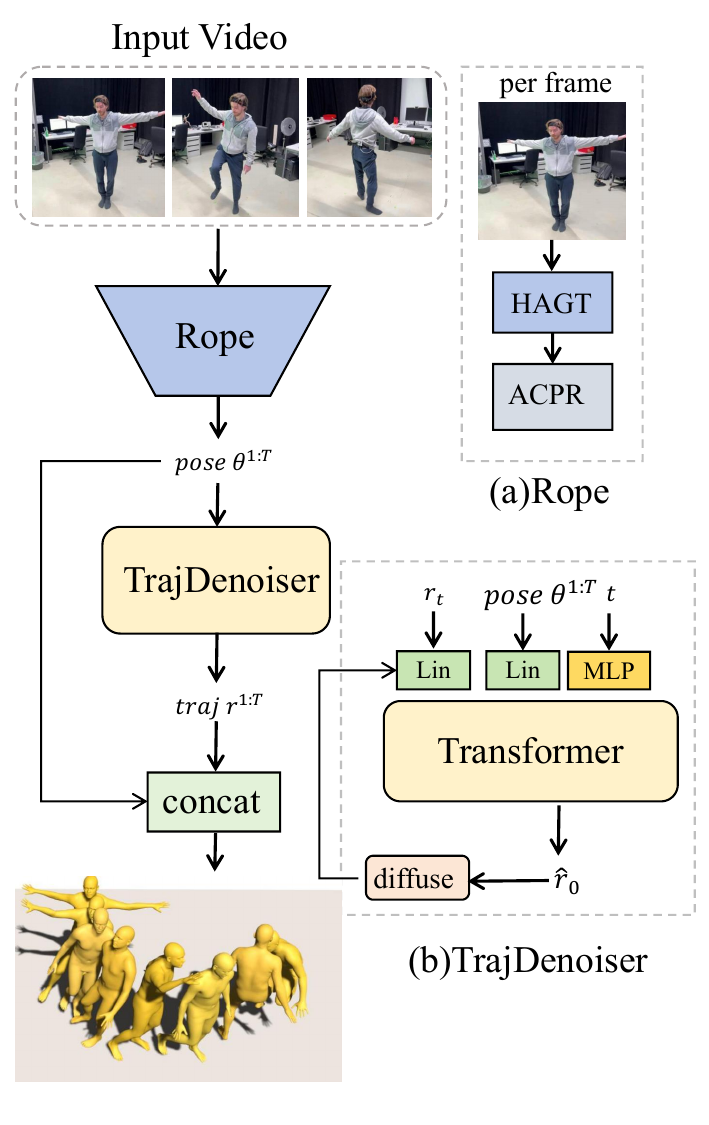}
  \end{center}
  \caption{
The overall structure of RopeTP. Here, Rope reconstructs the video results frame by frame. Meanwhile, TrajDenoiser regenerates the global trajectory of this sequence.}
\vspace{-2em}
  \label{fig:arc_full}
  \end{figure}

\subsection{Overview}
The overall structure of RopeTP is illustrated in Figure~\ref{fig:arc_full}. Its input is an RGB video captured by a monocular camera, and RopeTP's goal is to recover the global robust human mesh. To achieve this objective, we first employ the Rope module to estimate the precise human parameter sequence frame-by-frame, which serves as input to the SMPL model for obtaining 3D human joints $\theta \in \mathbb{R}^{N\times24\times3}$. Subsequently, we sample $x_{t} \sim \mathcal{N}(0,I)$ from pure Gaussian noise as the initial noise trajectory and linearly map local joint points to the feature space, which, along with the diffusion time steps, are fed into the TrajDenoiser. The regenerated global trajectory $\hat{r_{0}}\in \mathbb{R}^{N\times3}$ is ultimately obtained through 100 steps of DDIM denoising. The specific implementation details will be disclosed later in the manuscript.

\subsection{Network Architecture of Rope}
\begin{figure}[t]
\includegraphics[width=1.0\linewidth]{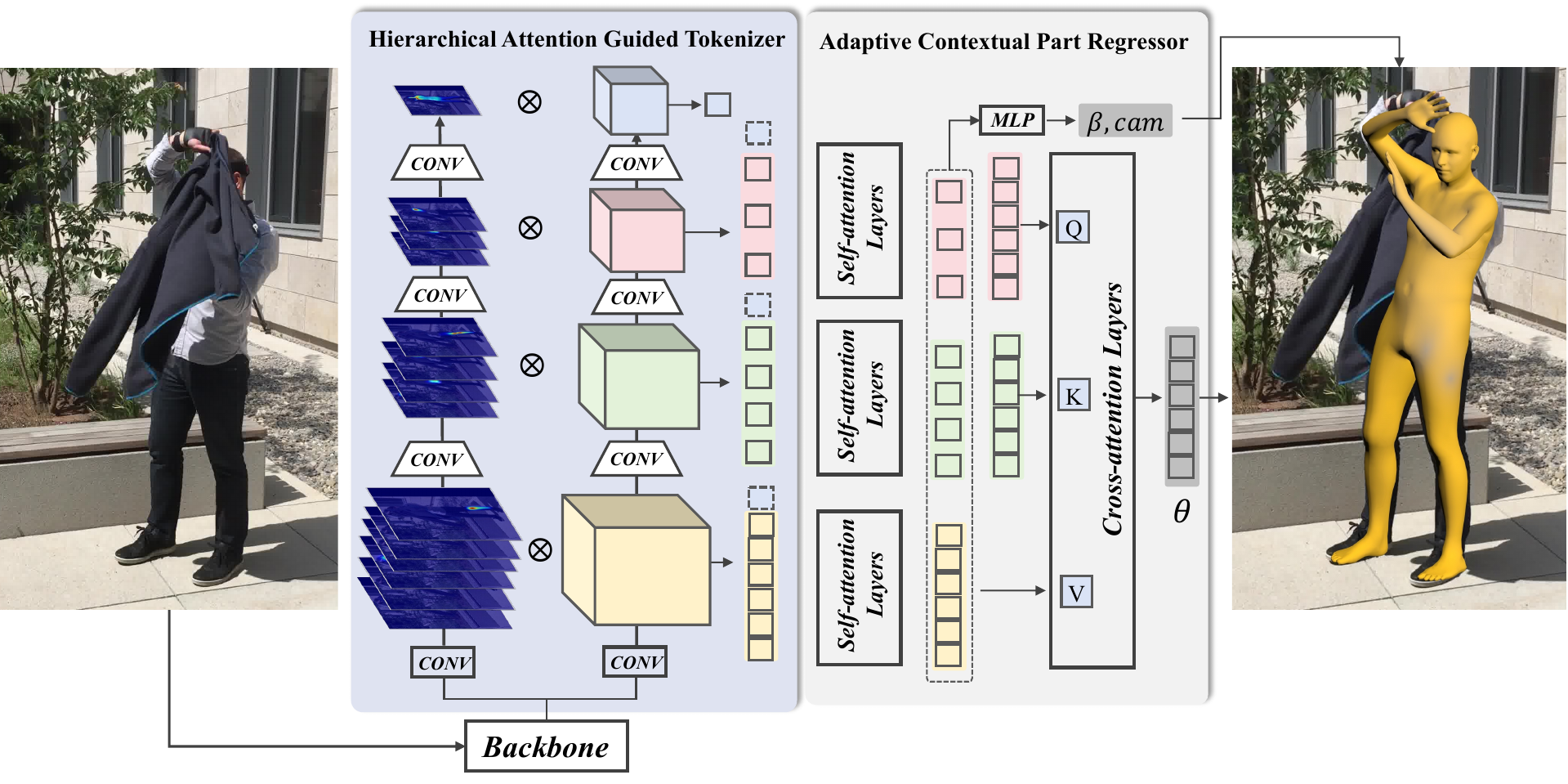}
    \captionof{figure}{ 
Overview of our \textbf{Rope} module. Given an RGB image as input, our method uses the Hierarchical Attention Guided Tokenizer to obtain multi-scale part and body features, combined into token sequences. The Adaptive Contextual Part Regressor optimizes scale-specific visual cues using a two-layer self-attention mechanism. Finally, we propose an efficient inter-hierarchical cross-attention layer to interact with visible information across scales and regress the SMPL parameters.}
  \label{fig:arc}
\end{figure}

Our Rope (\textbf{Ro}bust \textbf{P}ose \textbf{E}stimation) module draws inspiration from the human ability to infer missing information using contextual clues, much like solving a puzzle with some pieces hidden or missing. This approach is particularly relevant in the challenge of estimating human shape and pose from single-camera images, where occlusions are a common obstacle. In our method, we first define a hierarchical body division strategy that interpret ate human body parts into four levels. Then, as depicted in the Figure~\ref{fig:arc}, we employ the Hierarchical Attention Guided Tokenizer (HAGT) to extract body part features of four levels.\\
Since the human body exhibits different levels of coordination in various activities. By mirroring this inherent structure, we can capture the nuances of human movement more accurately. Moreover, different levels of body coordination provide varying visual cues, which are crucial in scenarios where parts of the body are occluded. Therefore, by dividing the body into hierarchical levels, we can utilize visible segments to infer the posture of occluded parts, akin to completing a puzzle using surrounding pieces.
As shown in Figure~\ref{fig:hbd}, we categorize the human body into four hierarchical levels of coordination, with a specific focus on handling occlusions:
\begin{figure}[t]
  \begin{center}
    \includegraphics[width=0.85\linewidth]{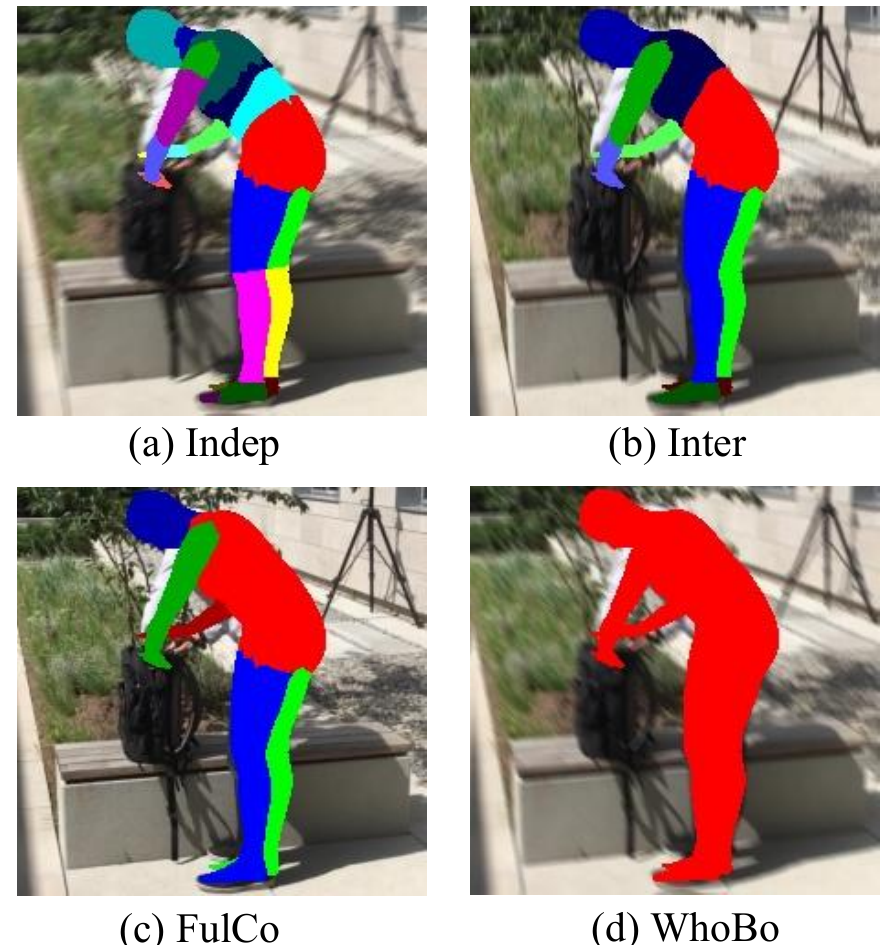}
  \end{center}
  \caption{\textbf{The visualization of body hierarchy levels.}}
  \vspace{-2em}
  \label{fig:hbd}
  \end{figure}

\noindent\textbf{Independent Level (Indep)}: \textit{Individual body parts like hands or feet and 24 parts in total, crucial for when specific areas are occluded.} \\
\noindent\textbf{Interdependent Level (Inter)}: \textit{Involves body parts that typically move in coordination, such as the forearm and upper arm, 11 parts in total, aiding in inferring occluded regions based on visible adjacent areas.}\\
\noindent\textbf{Fully Coordinated Level (FulCo)}: \textit{Encompasses larger body segments, like the entire arm, where complex movements occur, 6 parts in total, enabling the understanding of broader movement patterns even with partial visibility.}\\
\noindent\textbf{Whole Body Level (WhoBo)}: \textit{The entire body, integrating all levels for a holistic understanding of pose and movement.}\\
This hierarchical body division is fundamentally guided by the principle of capturing the varying degrees of motion coordination and complexity in human movements. This division is designed to reflect the natural way the human body operates and interacts with its environment, especially under occlusion conditions.

\noindent\textbf{Hierarchical Attention Guided Tokenizer.}
This module is designed to utilize the attention mechanism for extracting hierarchical part feature tokens. The attention mechanism is specifically fine-tuned to be sensitive to occlusions, which allows the framework to concentrate on available visual cues and effectively infer missing information. In our implementation, we pass the feature blocks obtained from the backbone through convolutional layers, resulting in attention maps at each coordinated hierarchy level, denoted as $Att_{\text{x}} \in \mathbb{R}^{H\times W\times J}$ (where x represents a specific scale level, and J stands for the number of parts at the corresponding scale level).
At the same time, to obtain part masks for explicit supervision, we render the ground truth (GT) labels as 3D human meshes and 3D/2D keypoints. Based on the weak-perspective projection strategy, we compute the projection matrix in reverse using 2D pixel coordinates and 3D coordinates. We then project the 3D human mesh onto the pixel space and classify it according to the mesh index numbers of each SMPL joint, resulting in GT masks at four scale levels. Additionally, we design another channel to extract 3D feature blocks, denoted as $V_{\text{x}} \in \mathbb{R}^{H\times W\times C}$ (where C represents the 3D feature channel dimension) from the features produced by the backbone network. Finally, we use the Hadamard operation to extract the ultimate multi-scale token sequences.
\begin{equation}
\begin{aligned}
  T_{\text{x}}=\sigma({{Att}_{\text{x}}})^{T} \odot V_{\text{x}}, 
\end{aligned}
\end{equation}
where $\odot$ is the Hadamard product and $\sigma(Att_{\text{x}})$ is used as a soft attention mask to aggregate features. This enables us to extract crucial information corresponding to each part, thereby facilitating the reconstruction of accurate results.

\noindent\textbf{Adaptive Contextual Part Regressor.}
To facilitate interaction with visual cues at various hierarchical levels and uncover appropriate reasoning relationships, we introduce the Adaptive Contextual Part Regressor (ACPR). Initially, it uses WhoBo features as reference dependencies and shares them across tokens of other hierarchical levels to create global-aware token sequences. The intra-hierarchical self-attention layers then adaptively learn part dependencies to optimize token sequences within each hierarchy level. Finally, we input the optimized sequences into cross-hierarchical attention layers to interact with visual cues across different hierarchical levels, reconstructing precise and robust mesh results based on a wealth of visible information.

\textbf{ISL} (\textbf{I}ntra-hierarchical \textbf{S}elf-Attention \textbf{L}ayer).
First, we expand the WhoBo features to connect them to part feature channels at each hierarchy level. Then, we construct three sets of token sequences at different hierarchy levels, $T_{\text{x}} \in R^{N\times C}$ (where N represents the number of queries corresponding to the hierarchy level, C denotes the fused feature dimension, and x refers to a specific scale level). Concurrently, we introduce a two-layer Multi-Head Self-Attention layer to encode global-aware part tokens $F'_{\text{x}} \in R^{N\times 128}$, enhancing visual cues within each scale level. As a result, our token sequences establish part dependencies within each hierarchy level, improving inference logic and endowing each token with the ability to perceive holistic information.

\textbf{ICL} (\textbf{I}nter-hierarchical \textbf{C}ross-Attention \textbf{L}ayer).
After obtaining the optimized token sequences at three hierarchical levels, we introduce the inter-hierarchical cross-attention layer to enable interaction of visual cues among tokens at different hierarchical levels. This allows features at all hierarchical levels to mutually reinforce each other, resulting in satisfactory and reasonable predictions. Specifically, we first expand the Inter and FulCo tokens to match the 24 joints of the SMPL model, such as the shoulder, arm, and elbow joints, which all correspond to the FulCo arm token. Next, diverging from the traditional cross-attention layer strategy that employs identical K and V values, we use token sequences from the three hierarchical levels as Q, K, and V matrices for the cross-attention layer input. This not only enhances computational efficiency but also more effectively interacts with visual cues across hierarchical levels, providing the model with a more robust feature selection capability. Furthermore, since the visual cues across hierarchical levels are highly redundant, necessary information is not lost while reducing computational demands (see \textbf{\textcolor{blue}{Supplementary Materials}} for details). The formula is as follows:
\begin{equation}
\begin{aligned}
  Attention_{\text{Indep}}&=\text{softmax}(\frac{Q_{\text{FulCo}}K_{\text{Inter}}^{T}}{\sqrt{d_{k}}})V_{\text{Indep}},\\
  Attention_{\text{Inter}}&=\text{softmax}(\frac{Q_{\text{Indep}}K_{\text{FulCo}}^{T}}{\sqrt{d_{k}}})V_{\text{Inter}},\\
  Attention_{\text{FulCo}}&=\text{softmax}(\frac{Q_{\text{Inter}}K_{\text{Indep}}^{T}}{\sqrt{d_{k}}})V_{\text{FulCo}},
\end{aligned}
\end{equation}
Subsequently, we associate three cross-hierarchical interaction visual cues and project them linearly to the SMPL rotation angles represented in rot6d through an MLP layer, serving as the final output. The camera and shape parameters are directly linked to the features of the three levels and are obtained through projection via an MLP layer.
\begin{equation}
\begin{aligned}
\text{Mix}&=\text{Concat}(Att_{\text{Indep}},Att_{\text{Inter}},Att_{\text{FulCo}})),\\
\theta&=MLP(LN(\text{Mix})),\\
\end{aligned}
\end{equation}
By inputting the regression-obtained rotation angles $\theta\in \mathbb{R}^{24\times6}$, camera parameters $cam\in \mathbb{R}^{3}$, and shape parameters $\beta\in \mathbb{R}^{10}$ into the SMPL model, we can acquire 3D mesh points $M_{3D}\in \mathbb{R}^{6890\times3}$ and 3D keypoints $J_{3D}\in \mathbb{R}^{24\times3}$. Furthermore, by utilizing camera parameters and weak perspective projection, we can also obtain 2D mesh points $M_{2D}\in \mathbb{R}^{6890\times2}$ and 2D keypoints $J_{2D}\in \mathbb{R}^{24\times2}$.
\subsection{Trajectory Denoiser}
In this chapter, we introduce a diffusion-based trajectory regeneration model. We represent the reconstructed human trajectory sequences from Rope as $r^{1:N}$ and associate them with the corresponding control signal $c$ (here, the encoded joint space positions p), where $N$ denotes the number of frames in the trajectory sequence. We employ a diffusion probability model~\cite{ho2020denoising} to regenerate human trajectories, where the diffusion model progressively anneals pure Gaussian noise into the trajectory distribution $p(r)$. Consequently, the model can predict noise from a T-time-step Markov noise process ${r_{t}^{1:N}}_{t}^{T}$, with $r_{0}^{1:N}$ directly sampled from the original data distribution. The diffusion process is as follows:
\begin{equation}
 q(r_{t}|r_{t-1}) = \mathcal{N}(r_{t};\sqrt{1-\beta_{t}}r_{t-1},\beta_{t}I), 
\end{equation}
\begin{equation}
 = \mathcal{N}(\sqrt{\frac{\alpha_{t}}{\alpha_{t-1}}}r_{t-1},(1-\frac{\alpha_{t}}{\alpha_{t-1}})I), 
\end{equation}
where $\{\beta_{t}\}_{t=1}^{T}$ is the variance schedule and $\alpha_{t} = \prod_{s=1}^{t}(1-\beta_{s})$. Then the reverse process becomes $p_{\theta}(r_{0:T}):=p(r_{T})\prod_{t=1}^{T}p_{\theta}(r_{t-1}|r_{t})$, starting from $r_{T}\sim\mathcal{N}(0,I)$ with noise predictor $\epsilon_{t}^{\theta}$:
\begin{equation}
 r_{t-1}= \frac{1}{\sqrt{1-\beta_{t}}}(r_{t}-\frac{\beta_{t}}{\sqrt{1-\alpha_{t}}}\epsilon_{t}^{\theta}(r_{t}))+\sigma_{t}z_{t},\\
\end{equation}
where $z_{t} \sim \mathcal{N}(0,I)$ and $\sigma_{t}^{2}=\beta_{t}$ means the variance schedule stays constant.\\
However, unlike the original DDPM~\cite{ho2020denoising}, we consider the inherent extreme physical constraints of 3D human bodies and replace the predicted noise with the original human trajectory, deviating from image generation. As illustrated in Figure~\ref{fig:arc_full}, at each step of the denoising process, we reconstruct the original trajectory from pure Gaussian noise, ultimately producing the final generated result through a cyclic process of noise addition and denoising:
\begin{equation}
    \hat{r}_{0}= \frac{r_{t}-\sqrt{1-\alpha_{t}} \epsilon_{t}^{\theta}\left(r_{t}|c\right)}{\sqrt{\alpha_{t}}},
\end{equation}
\begin{equation}
    r_{t-1}=\sqrt{\alpha_{t-1}} \hat{r}_{0}+\sqrt{1-\alpha_{t-1}-\sigma_{t}^{2}} \cdot \epsilon_{t}^{\theta}\left(r_{t}|c\right)+\sigma_{t} z_{t},
\end{equation}
where $c$ represents the joint space positions after being encoded by the conditional signal encoder and $\sigma$ represents a unified form of the non-Markovian process, where we consistently maintain the DDIM process with a constant value of 0. The regenerated trajectory, combined with joint space coordinates, undergoes optimization in a physics simulation environment~\cite{yuan2023learning}, resulting in our final output.

\subsection{Loss Function}
As Rope module is an end-to-end process trained through explicit supervision, our objective is to minimize the discrepancy between the predicted results and the ground truth labels as much as possible. To capture the prediction errors, we define the following loss functions for our network:
\begin{equation}
\begin{aligned}
    Loss&=L_{\text{SMPL}}+L_{\text{3D}}+L_{\text{2D}}+L_{\text{Att}},\\
L_{\text{3D}}&=||J_{\text{3D}}-\hat{J_{\text{3D}}}||_{F}^{2},\\
L_{\text{2D}}&=||J_{\text{2D}}-\hat{J_{\text{2D}}}||_{F}^{2},\\
L_{\text{SMPL}}&=||\theta-\hat{\theta}||_{2}^{2} + ||\beta-\hat{\beta}||_{2}^{2},\\
L_{\text{Att}}&=\!\frac{1}{HW}\!\sum\limits_{h,w}\limits^{}\!CrossEntropy\!(\!\sigma(Att_{h,w})\!,\!{Seg_{h,w}}\!),
\end{aligned}
\end{equation}
where $\hat{X}$ represents the ground truth for the corresponding variable $X$. $Att_{h,w}$ represents the predicted 2D attention map within the HAGT module, while $Seg_{h,w}$ refers to the 2D segmentation map mentioned in the Supplementary Material, with the latter serving as the supervision label for the former. We train the model with all losses for the first 100 epochs. After that, we remove the segmentation supervision loss and continue training the model. This initial training phase helps the model learn to recognize and emphasize relevant regions in the input images. Once the attention mechanism has been guided towards the body parts, we remove the segmentation supervision loss and continue training. During this phase, the attention mechanism adapts further to better recover the human mesh from the images.

Our trajectory denoiser consists of a Transformer-based denoiser and a condition control signal encoder. The former is composed of 12 layers and 8-head self-attention modules with default skip connections, while the latter consists of Linear layers. The Linear Block maps joint coordinates to generate control conditions and linearly maps them to the same dimensional space as the noise. We then concatenate the two as input for the denoiser. For the training of the trajectory denoiser module, we use a single A100 GPU with a batch size of 256 and a learning rate set to 3e-5. To ensure trajectory accuracy, we disable the CFG module and rely solely on condition control guidance during the denoising process. During training, we adopt the following reconstruction loss and physical constraints:
\begin{equation}
\begin{aligned}
    \mathcal{L}_{simple}&=E_{x_{0}~q(x_{0}|c),t~[1,T]}[HuberLoss(x_{0}-\hat{x}_{0})],\\
    \mathcal{L}_{foot}&=\frac{1}{N-1}\sum_{i=1}^{N-1}\|\hat{x}_{0}^{i+1}-\hat{x}_{0}^{i} \| f_{i},\\
\end{aligned}
\end{equation}
where $f_{i}$ determines the ground contact state by calculating the rate of change in the y-axis position of the footstep. 

\begin{table}
\resizebox{\linewidth}{!}{
  \centering
  
  \begin{tabular}{l|l|ccc ccc ccc cc}
    \toprule[1.5pt]
    \multirow{2}{*}{Method} & \multirow{2}{*}{Source}& \multicolumn{3}{c}{3DPW\cite{von2018recovering} }  & 

    \multicolumn{2}{c}{Human3.6M \cite{ionescu2013human3}} \\
    
      & & MPJPE$\downarrow$ & PA-MPJPE$\downarrow$ & MPVPE$\downarrow$ & MPJPE$\downarrow$ &  PA-MPJPE$\downarrow$ \\
    \midrule[0.75pt]
\textbf{RopeTP} (HRNet-48) & Ours & \textbf{\textcolor{red}{65.2}} & \textbf{\textcolor{blue}{42.3}} & \textbf{\textcolor{red}{78.0}}    &    \textbf{\textcolor{blue}{46.8}}  &  \textbf{\textcolor{red}{32.2}}     \\
\textbf{RopeTP} (HRNet-32) & Ours & 66.8 & 44.3 & 80.7    &    47.4  &  34.1     \\
\midrule[0.75pt]    
MotionBert (HRNet-48)~\cite{zhu2022motionbert}        & ICCV2023 & 68.8   & \textbf{\textcolor{red}{40.6}} & \textbf{\textcolor{blue}{79.4}} & - & - \\
HMR2.0 (ViT)~\cite{goel2023humans} &ICCV2023 &70.0 &44.5 &- &\textbf{\textcolor{red}{44.8}} &33.6 \\
PLIKS(HRNet-32)~\cite{shetty2023pliks} &CVPR2023 &\textbf{\textcolor{blue}{66.9}} &42.8 &82.9 &49.3 &34.7\\
NIKI(HRNet-48)~\cite{li2023neural}&CVPR2023 &71.3 &40.6 &86.6 &- &- \\
IKOL(ResNet-34)~\cite{zhang2023ikol}&AAAI2023 &71.7 &84.1 &44.5 &- &-\\
CLIFF(HRNet-48)~\cite{li2022cliff} &ECCV2022 & 69.0 & 43.0 & 81.2 & 47.1 & \textbf{\textcolor{blue}{32.7}}\\
VisDB(HRNet-32)~\cite{yao2022learning} &ECCV2022 & 72.1 & 44.1 & 83.5 & - & -\\
PARE(HRNet-32)~\cite{kocabas2021pare}&ICCV2021 &74.5 &46.5 &88.6 &- &- \\
METRO(HRNet-64)~\cite{lin2021end}&ICCV2021 &77.1&47.9&88.2 &54.0 &36.7 \\
ROMP(HRNet-32)~\cite{sun2021monocular}&ICCV2021 &82.7 &60.5 &- &- &- \\
HybrIK(HRNet-32)~\cite{li2021hybrik}&CVPR2021 &80.0 &48.8 &94.5 &54.4 &34.5\\

    \bottomrule[1.5pt]
  \end{tabular}}
\caption{Our approach is compared with the latest results using two benchmark datasets, including 3DPW and Human3.6M. 3DPW is a highly challenging outdoor dataset, while Human3.6M is a high-quality indoor dataset. The best and second-best results are highlighted in \textcolor{red}{red} and \textcolor{blue}{blue}, respectively.}
  \label{tab:main_tab}

\end{table}

\section{Experiments}

Here, we initially conduct a comparative analysis with other advanced methods on two publicly available generic 3D human pose datasets. Subsequently, we provide comprehensive ablation studies to validate the contribution of each module and the effectiveness of our structural design, thereby highlighting the robustness and superiority of our approach.

\subsection{Comparison of 3D Human Reconstruction}
We initially conduct a comprehensive comparison with other methods on two publicly available benchmarks, namely 3DPW~\cite{von2018recovering} and Human3.6M~\cite{ionescu2013human3}. For the 3DPW dataset, we start by training on a combined dataset (COCO, Human3.6M, MPII, and 3DHP) and then fine-tune the network on the 3DPW training dataset. In contrast, for Human3.6M, we exclusively train on the mixed dataset and evaluate according to the official protocol 2. Additionally, we present two network configurations, one with and one without global-aware references from WhoBo, and showcase the corresponding results. Table~\ref{tab:main_tab} displays the comparative results of various approaches.

Our method outperforms nearly all existing techniques in terms of MPJPE and MPVPE. Particularly noteworthy is the significant improvement in MPVPE, an indicator that our reconstructed results more accurately conform to the human body in the original image. Unlike HMR2.0, which is entirely based on a Transformer structure, our method manages to maintain a reconstruction speed close to that of baseline models, PARE and CLIFF, while reducing MPVPE from 81.2 to 78.0, even without a global perception reference (from 81.2 to 80.2). Considering that the latest methods, such as NIKI and PLIKS, have not released open-source training and evaluation models, we compare our approach based on the original performance metrics. Although these new methods exhibit superior performance on the PA-MPJPE metric in the 3DPW dataset, it is important to note that they utilize additional modules (e.g., 3D keypoint priors, posterior distribution sampling). In contrast, our regression-based end-to-end approach is more straightforward and easier to train. HMR2.0, on the other hand, is constrained by its pure ViT architecture, resulting in a slow regression speed that cannot support real-time tasks in practical applications. Supported by subsequent qualitative analysis, the experimental results validate that our approach not only predicts more accurate mesh and body posture, but also exhibits robustness against the instability caused by occlusions.

\begin{table}

\resizebox{1\linewidth}{!}{
  \centering
  
  \begin{tabular}{l|l|cc cc}
    \toprule[1.5pt]
    \multirow{2}{*}{Method} & \multirow{2}{*}{Source}& \multicolumn{2}{c}{3DPW-OCC\cite{von2018recovering} }  & \multicolumn{2}{c}{3DOH\cite{zhang2020object} } \\
    
      & & MPJPE$\downarrow$  & PA-MPJPE$\downarrow$ & MPJPE$\downarrow$  & PA-MPJPE$\downarrow$ \\
    \midrule[0.75pt]

\textbf{RopeTP} & Ours & \textbf{79.7} & \textbf{48.7} & \textbf{77.8} & \textbf{53.7}\\
\midrule[0.75pt]  
CLIFF~\cite{kocabas2020vibe} &ECCV2022 &82.7 &51.9 &80.4 &58.7\\
PARE~\cite{choi2021beyond} &ICCV2021 &84.9 &57.5 &94.9 &62.4\\

    \bottomrule[1.5pt]
  \end{tabular}}
\caption{Performance comparison on occlusion datasets 3DPW-OCC and 3DOH. Bold indicates best results.}
  \label{tab:occ}

\end{table}
\subsection{Occlusion Robustness}
Table~\ref{tab:occ} presents the evaluation results on the occlusion datasets 3DPW-OCC and 3DOH~\cite{zhang2020object} for our proposed method, the baseline PARE, CLIFF, and other regression-based methods like BoPR. Specifically, 3DPW-OCC is an occluded subset of the 3DPW dataset, while 3DOH is a challenging dataset proposed by Zhang et al., featuring extensive human body occlusions. All methods are evaluated on 3DPW-OCC without using the 3DPW training dataset. During the 3DOH evaluation, the network is fine-tuned on the 3DPW training set. Both evaluations exclude their respective training sets to ensure a fair comparison of generalization ability on occlusion datasets. PARE and CLIFF are retrained by us, while BoPR is evaluated using the original metrics. Our method outperforms previous ones in terms of MPJPE and PA-MPJPE metrics. Specifically, on the 3DPW-OCC dataset, our method reduces the errors of both metrics by 4.8\% and 6.6\% respectively compared to the baseline. These quantitative results demonstrate that cross-level attention inference effectively alleviates occlusion challenges, enhancing accuracy and generalization ability in demanding scenarios.

\begin{figure}
  \begin{center}
    \includegraphics[width=1.0\linewidth]{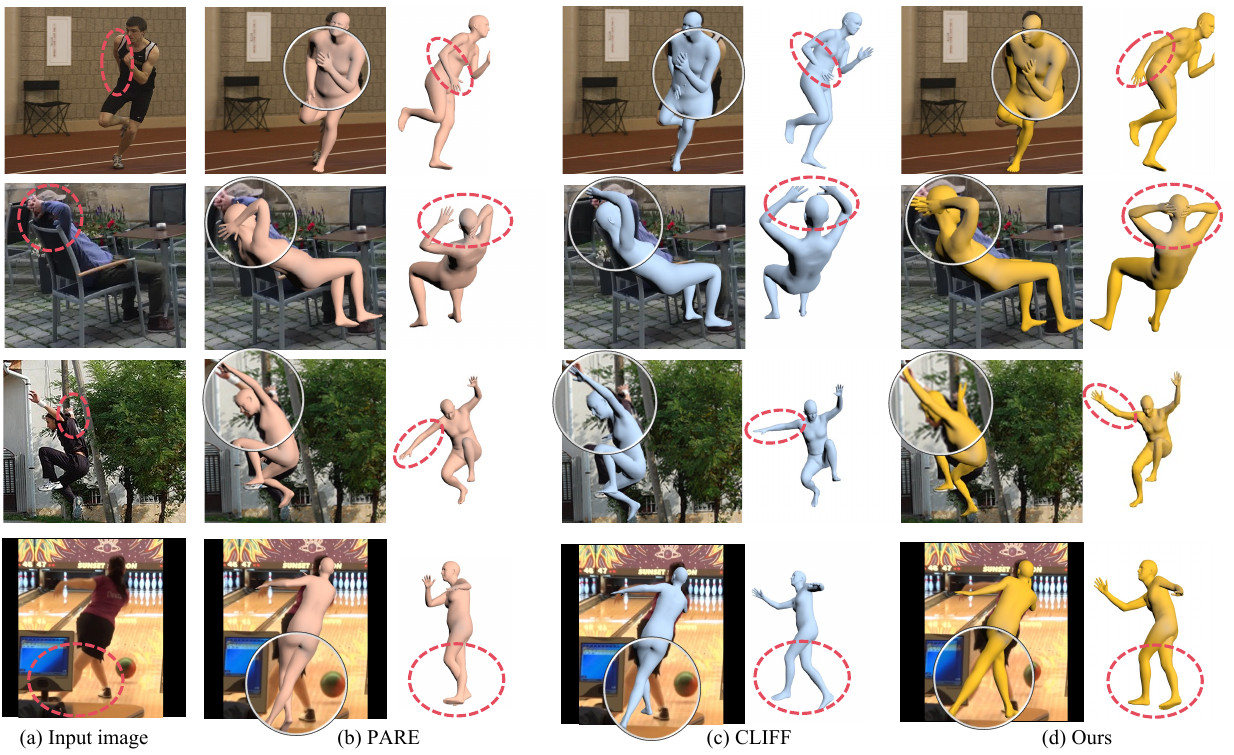}
  \end{center}
  \caption{\textbf{Qualitative comparison on 3DPW and in-the-wild datasets.} (a)
Input images. (b) Results by PARE~\cite{kocabas2021pare}. (c) Results by CLIFF~\cite{li2022cliff}. (d) Results by Ours.}
  \label{fig:quat}
  \end{figure}

In Figure~\ref{fig:quat}, we demonstrate the performance of three methods (our proposed approach, PARE, and CLIFF) on the 3DPW test set and outdoor datasets under occlusion conditions. PARE represents a local regression method based on a single attention-guiding mechanism, while CLIFF is a single-output method based on global feature regression. The results indicate that our approach effectively reconstructs human body meshes in both indoor and outdoor real-world scenarios. Notably, under the occlusion conditions shown in the third row, previous methods struggle to reasonably infer occluded limbs. Relying solely on visible right wrist information, both PARE and CLIFF incorrectly infer the self-occluded right arm as a naturally hanging posture. In contrast, our approach accurately reconstructs the upwardly extended arm posture through a cross-hierarchical inference strategy. Moreover, our approach demonstrates a significant advantage in fitting the mesh to the original image compared to previous methods. This qualitative analysis emphasizes the superiority of the multi-level attention inference approach in addressing challenging occlusion scenarios.

\subsection{Diffusion-Based Trajectory Regeneration}
\begin{figure}
  \begin{center}
    \includegraphics[width=1.0\linewidth]{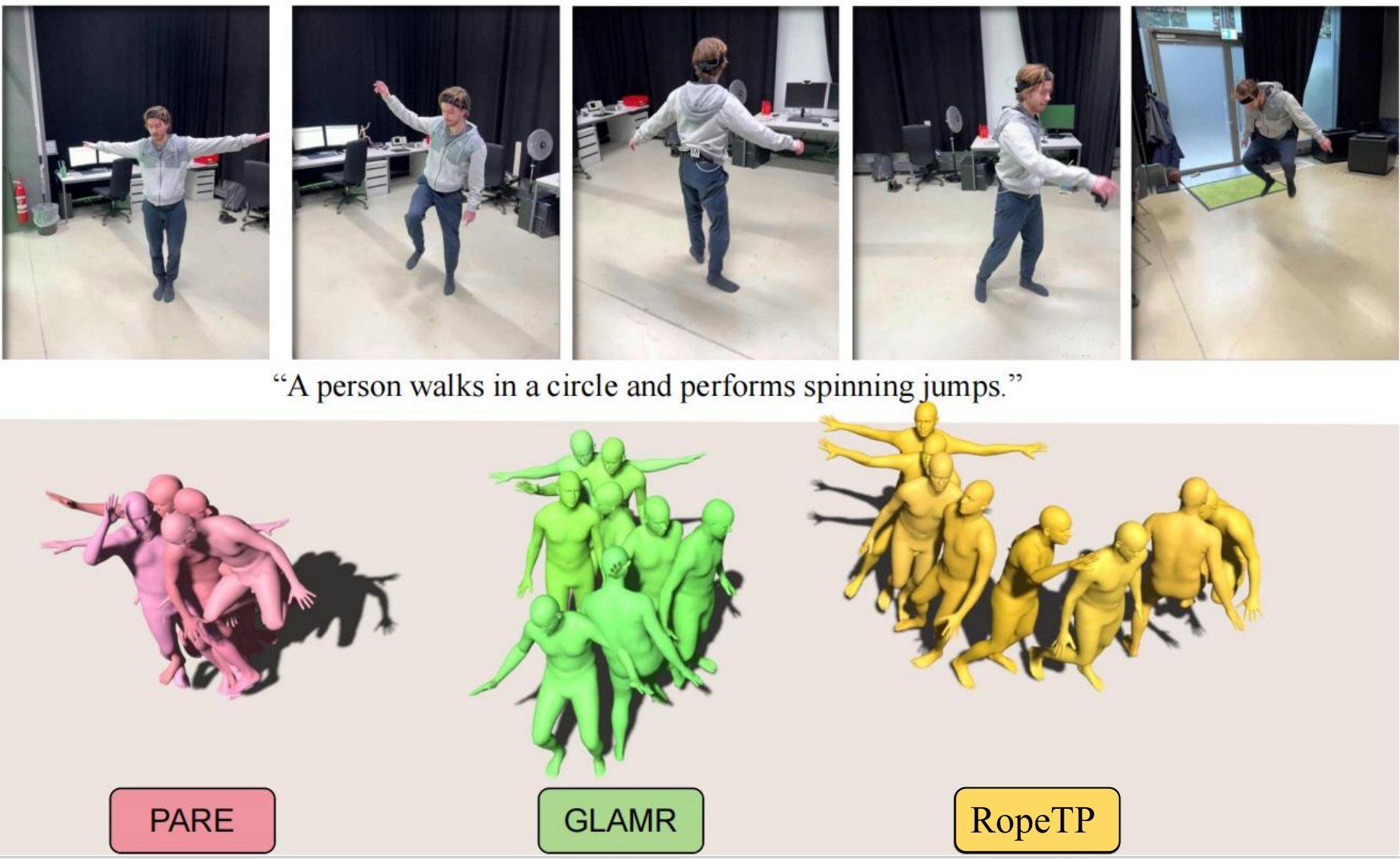}
  \end{center}
  \caption{Global qualitative comparison on the EMDB dataset.} 
  \label{fig:quat_dt}
  \end{figure}

To evaluate the accuracy of our regenerated global trajectories, we introduce an additional EMDB dataset~\cite{kaufmann2023emdb}, which provides real global motion through a dynamic camera. We substitute the global trajectory of the original Rope with the trajectory regenerated based on the diffusion model, which we refer to as RopeTP, and compare it with state-of-the-art methods.  Evaluation in the single-frame camera coordinate system is sorted in descending order by MPJPE. We use the EMDB1 sequence and 24 joint count for evaluation under this protocol. W-MPJPE and WA-MPJPE are employed for global motion accuracy assessment on EMDB2. As shown in the table~\ref{tab:dtt}, RopeTP outperforms existing methods on all metrics. Specifically, under single-frame comparisons in the camera coordinate system, both PARE and GLAMR fall short of our RopeTP. In the world coordinate system, RopeTP's global spatial error is superior to GLAMR and TRACE. We further substantiate this in the figure~\ref{fig:quat_dt}. As shown, RopeTP significantly outperforms GLAMR in capturing human motion patterns in the global coordinate system, while PARE remains almost stationary near the origin. Our method's superior performance can be attributed to the effective combination of the Rope module and the diffusion-based trajectory regeneration, which provides a more comprehensive understanding of human motion and results in better global trajectory estimations. Furthermore, the global comparison results between RopeTP and PARE indicate that accurate human poses as trajectory priors better facilitate the recovery of global robust human forms.

\begin{table}

\resizebox{1\linewidth}{!}{
  \centering
  
  \begin{tabular}{lccccccc}
    \toprule[1.5pt]
    \multirow{2}{*}{Method} & \multicolumn{3}{c}{EMDB1(24)} & \multicolumn{2}{c}{EMDB2} \\ 
 & MPJPE$\downarrow$ & PAMJE$\downarrow$ & MPVE$\downarrow$  &W-MPJPE$\downarrow$ & WA-MPJPE$\downarrow$ \\ 
    
    \midrule[0.75pt]
HyBrIK~\cite{li2021hybrik} &103.0 &\textbf{65.6} &122.2 &- &- \\ 
PARE~\cite{kocabas2021pare} &113.9 &72.2 &133.2 &914.3 &434.0 \\ 
CLIFF~\cite{li2022cliff} &103.3&68.1 &128.0 &- &- \\
TRACE~\cite{sun2023trace}   &109.9 &70.9 &127.4 &2244.9 &544.1\\
\textbf{RopeTP (Ours)} &99.1 &65.8 &\textbf{110.5} &\textbf{622.7} &\textbf{260.9}\\
\hline

SLAHMR~\cite{ye2023decoupling} &\textbf{93.5}  &69.5 &110.7 &807.4 &336.9\\
GLAMR~\cite{yuan2022glamr} &113.6  &73.5 &133.4 &756.1 &286.2\\

    \bottomrule[1.5pt]
  \end{tabular}}
\caption{Quantitative comparison of state-of-the-art models on the EMDB dataset. Bold values indicate the best results. The global results with PARE as the trajectory prior were tested on EMDB2.}
  \label{tab:dtt}
\end{table}

\begin{table}

\resizebox{1\linewidth}{!}{
  \centering
  
  \begin{tabular}{lccccccc}
    \toprule[1.5pt]
    \multirow{2}{*}{Method} & \multicolumn{2}{c}{3DPW-TEST} & \multicolumn{2}{c}{3DPW-OCC} \\ 
 & MPJPE$\downarrow$ & MPVPE$\downarrow$ & MPJPE$\downarrow$ & PAMJE$\downarrow$ \\ 
    
    \midrule[0.75pt]

Baseline~\cite{kocabas2021pare} &74.5 &88.6 &84.9 &57.5 \\ 
Baseline+TF &73.2  &85.0 &84.6 &55.8\\ 
Baseline+HAGT+MLP   &72.0 &83.4 &83.1 &53.0\\\hline
\textbf{Rope(Ours)} w/o WhoBo &68.3  &81.5 &80.9 &49.9\\
\textbf{Rope(Ours)} &66.8 &80.7 &79.7 &48.7\\

    \bottomrule[1.5pt]
  \end{tabular}}
\caption{Ablation study on \textbf{Rope}. Among them, TF means Transformer, and all methods fine-tune the network on the 3DPW training set when evaluating on the 3DPW-TEST. 'w/o' means without this module.}
  \label{tab:abl1}
\end{table}
\subsection{Ablation Study}
Ablation study on \textbf{Rope}, we conduct ablation studies on the 3DPW-TEST and 3DPW-OCC datasets to thoroughly evaluate the impact of various components within our method. We present the results in terms of MPJPE and MPVPE metrics. Table~\ref{tab:abl1} outlines our ablation results, focusing on the significance of the Hierarchical Attention Guided Tokenizer (HAGT) and Adaptive Context Part Regressor (ACPR) modules in handling occlusions. Initially, we incorporate the Transformer structure into the baseline model PARE, resulting in limited improvement due to constrained visual cues at the Indep level. Next, we introduce HAGT to capture multi-level visible information and achieve a substantial enhancement in the MPVPE metric through simple fusion, significantly mitigating occlusion interference. Ultimately, we eliminate the reliance on WhoBo reference information and depend exclusively on the interaction of visual cues from three distinct hierarchical levels for inference. This approach still marginally surpasses the BoPR, which depends on reference features. Furthermore, this cross-hierarchical visual information interaction substantially strengthens the model's robustness against occlusions, leading to a remarkable improvement on the 3DPW-OCC dataset.
\begin{table}
\resizebox{1.0\linewidth}{!}{
  \centering
  
  \begin{tabular}{lccccccc}
    \toprule[1.5pt]
    \textbf{\textcolor{red}{Table A}} & MPJPE$\downarrow$ & PAMJE$\downarrow$ & MPVE$\downarrow$  &W-MPJPE$\downarrow$ & WA-MPJPE$\downarrow$\\ 

    \midrule[0.75pt]
PARE &113.9 &72.2 &133.2 &13192.85 &15914.17\\ 
GLAMR &113.6  &73.5 &133.4 &756.1 &286.2 \\ 
\textbf{Rope (Ours)} &\textbf{93.7} & \textbf{63.0} &114.2 & 12058.74 &10896.93 \\
\hline
PARE+TP &113.9 &72.2 &133.2 &914.3 &434.0\\
Indep+TP &103.42 &66.81 &126.7 &752.71 &330.5\\
Inter+TP  &101.2 & 69.3 &122.54 &743.8 &309.2\\
Fulco+TP &98.2 &65.76 &120.82 &826.21 &368.0\\
\hline
RopeTP(rot6d con) &\textbf{93.7} & \textbf{63.0} &114.2 &778.25 &402.32\\
RopeTP(loc rep) &\textbf{93.7} & \textbf{63.0} &114.2 &711.3 &320.0\\
RopeTP w/o imitation &\textbf{93.7} & \textbf{63.0} &114.2 & 692.91 &284.27\\ 
\textbf{RopeTP (Ours)} & 99.1 & 65.8 &\textbf{110.5} &\textbf{622.7} &\textbf{260.9}\\
    \bottomrule[1.5pt]
  \end{tabular}}
  \caption{Ablation study on \textbf{TP}. 'w/o' means without this module.}
  \vspace{-10pt}
  \label{tab:dtt}
\end{table}

Ablation study on \textbf{TP}, Table~\ref{tab:dtt} presents the EMDB evaluation values at each scale level after trajectory generation using the Trajectory Predictor (TP). The results indicate that large-scale features effectively handle occlusion but suffer from larger global displacement errors due to inaccurate end positions. In contrast, small-scale features excel in global metrics but struggle with occlusion (worse MJE).

Furthermore, we performed additional ablation on the TP module's representation. Our approach, which uses joint location as the guiding condition and regenerates xyz coordinates, outperforms other methods (rot6d representation). The term "w/o imitation" refers to the absence of physical simulation optimization. Simulated gravity and friction forces significantly reduce issues like sliding and floating, enhancing global metrics. However, the built-in IK mechanism adjusts global actions, making them subjectively more natural but with a minor impact on metrics.\\
\section{Conclusion}
In summary, our innovative method addresses occlusion challenges in human body mesh reconstruction by introducing a trajectory diffusion generation model for optimizing monocularly-captured motion trajectories. Adaptable to diverse 3D applications, the Rope module employs hierarchical attention-guided segmenters and adaptive context part regressors to capture multi-level visual cues for accurate mesh inference under severe occlusion. Our method outperforms state-of-the-art approaches, particularly in heavily occluded scenes, demonstrating its potential for robust pose estimation and 3D reconstruction. Furthermore, RopeTP regenerates trajectories using a diffusion model, enhancing its applicability to motion capture multimedia applications (refer to \textbf{Supplementary Video}). However, some rough physical problems can also be implemented by referring to knowledge in the field of reinforcement learning and generation~\cite{yang2024freetalker, cheng2024expgest, cheng2024conditional, han2024reindiffuse, yu2023signavatars, han2024hutumotion}. At the same time, we can also use some detection methods~\cite{sun2024bidirectional} to evaluate the posture parameters of hands and faces separately to achieve global human body capture.
{\small
\bibliographystyle{ieee_fullname}
\bibliography{egbib}
}

\end{document}